\documentclass[10pt,twocolumn,letterpaper]{article}

\usepackage{cvpr}
\usepackage{times}
\usepackage{epsfig}
\usepackage{graphicx}
\usepackage{amsmath}
\usepackage{amssymb}
\usepackage{cuted}
\usepackage{caption}
\usepackage{subcaption}
\usepackage{bm}
\usepackage{xcolor}
\usepackage{flushend}
\usepackage{wrapfig}
\usepackage{pdfpages}

\newcommand{\bF}{{\bf F}}
\newcommand{\bH}{{\bf H}}
\newcommand{\bT}{{\bf T}}
\newcommand{\bB}{{\bf B}}
\newcommand{\balpha}{{\bm \alpha}}
\newcommand{\bbeta}{{\bm \beta}}
\newcommand{\bdelta}{{\bm \delta}}
\newcommand{\bgamma}{{\bm \gamma}}

\newcommand{\bp}{{\bf p}}
\newcommand{\bR}{{\bf R}}
\newcommand{\bt}{{\bf t}}

\newcommand{\depthmap}{{d}}
\newcommand{\paravspace}{\vspace{-10pt}}
\newcommand{\eqvspace}{\vspace{-0pt}}

\DeclareMathOperator{\E}{\mathbb{E}}

\graphicspath{{figures/}}

\usepackage[pagebackref=true,breaklinks=true,letterpaper=true,colorlinks,bookmarks=false]{hyperref}

\cvprfinalcopy 

\ifcvprfinal\fi

\begin{document}
	
	\title{Deep 3D Portrait from a Single Image}
	
	\author{Sicheng Xu$^{1}$\thanks{This work was done when S. Xu was an intern at MSRA.} \ \ \, Jiaolong Yang$^{2}$ \ \ Dong Chen$^{2}$ \ \ Fang Wen$^{2}$ \ \ Yu Deng$^{3}$ \ \ Yunde Jia$^{1}$ \ \ Xin Tong$^{2}$ \\
		$^1${Beijing Institute of Technology} \quad
		$^2${Microsoft Research Asia} \quad $^3${Tsinghua University} \\
	}
	
	\maketitle
	\begin{abstract}
		In this paper, we present a learning-based approach for recovering the 3D geometry of human head from a single portrait image. Our method is learned in an unsupervised manner without any ground-truth 3D data. 
		We represent the head geometry with a parametric 3D face model together with a depth map for other head regions including hair and ear. A two-step geometry learning scheme is proposed to learn 3D head reconstruction from in-the-wild face images, where we first learn face shape on single images using self-reconstruction and then learn hair and ear geometry using pairs of images in a stereo-matching fashion. The second step is based on the output of the first to not only improve the accuracy but also ensure the consistency of overall head geometry. 
		We evaluate the accuracy of our method both in 3D and with pose manipulation tasks on 2D images. We alter pose based on the recovered geometry and apply a refinement network trained with adversarial learning to ameliorate the reprojected images and translate them to the real image domain. Extensive evaluations and comparison with previous methods show that our new method can produce high-fidelity 3D head geometry and head pose manipulation results.
		\vspace{-3pt}
	\end{abstract}
	
	\section{Introduction}
	Reconstructing 3D face geometry from 2D images has been a longstanding problem in computer vision. Obtaining full head geometry will enable more applications in games and virtual reality as it provides not only a new way of 3D content creation but also image-based 3D head rotation (\ie, pose manipulation). Recently, single-image 3D face reconstruction has seen remarkable progress with the enormous growth of deep convolutional neutral networks (CNN)~\cite{tewari2017mofa,tran2017regressing,guo2018cnn,genova2018unsupervised,deng2019accurate}.
	However, most existing techniques are limited to the facial region reconstruction without addressing other head regions such as hair and ear. 
	
	Face image synthesis has also achieved rapid progress with deep learning.
	However, few methods can deal with head pose manipulation from a singe image, which necessitates substantial image content regeneration in the head region and beyond.
	Promising results have been shown for face rotation~\cite{yin2017towards,bao2018towards,hu2018pose} with generative adversarial nets (GAN), but generating the whole head region with new poses is still far from being solved. One reason could be implicitly learning the complex 3D geometry of a large variety of hair styles and interpret them onto 2D pixel grid is still prohibitively challenging for GANs.
	
	In this paper, we investigate explicit 3D geometry recovery of  portrait images for head regions including face, hair, and ear.
	We model 3D head with two components: a face mesh by the 3D Morphable Model (3DMM)~\cite{blanz1999morphable}, and a depth map for other parts including hair, ear, and other regions not covered by the 3DMM face mesh.
	The 3DMM face representation facilitates easy shape manipulation given its parametric nature, and depth map provides a convenient yet powerful representation to model complex hair geometry.
	
	Learning single-image 3D head geometry reconstruction is a challenging task.
	At least two challenges need to be addressed here. 
	First, portrait images come with ground-truth 3D geometry are too scarce for CNN training, especially for hair which can be problematic for 3D scanning. To tackle this issue, we propose an unsupervised learning pipeline for head geometry estimation. For face part, we simply follow recent 3D face reconstruction methods~\cite{tewari2017mofa,genova2018unsupervised,godard2017unsupervised,zhou2017unsupervised} to learn to regress 3DMM parameters on a corpus of images via minimizing the rendering-raw input discrepancy. But for hair and ear, we propose to exploit view change and train on pairs of portrait images extracted from videos via minimizing appearance reprojection error. The second challenge is how to ensure a consistent head structure since it consists of two independent components. 
	We propose a two-step shape learning scheme where we use the recovered face geometry as conditional input of the depth network, and the designed loss function considers the layer consistency between face and hair geometry. We show that our two-step unsupervised shape learning scheme leads to compelling 3D head reconstruction results.
	
	Our method can be applied for portrait image head pose manipulation, the quality of which will be contingent upon the 3D reconstruction accuracy thus could be used to evaluate our method. Specifically, we change the pose of the reconstructed head in 3D and reproject it onto 2D image plane to obtain pose manipulation results. The reprojected images require further processing. Notably, the pose changes give rise to missing regions that need to be hallucinated. To this end, we train a refinement network using both real unpaired data and synthetic paired data generated via image corruption, together with a discriminator network imposing adversarial learning. Our task here appears similar to image inpainting. However, we found the popular output formation scheme (raw image merged with network-generated missing region) in deep generative image inpainting~\cite{yu2018generative,nazeri2019edgeconnect} leads to inferior results with obvious artifacts. We instead opt for regenerating the whole image. 
	
	Our contributions can be summarized as follows:
	\vspace{-4pt}
	\begin{itemize}
		\item  We propose a novel unsupervised head geometry learning pipeline without using any ground-truth 3D data. The proposed two-step learning scheme yields consistent face-hair geometry and compelling 3D head reconstruction results.
		\vspace{-3pt}
		\item We propose a novel single-image head pose manipulation method which seamlessly combines learned 3D head geometry and deep image synthesis. Our method is fully CNN-based, without need for any optimization or postprocessing.
		\vspace{-3pt}
		\item We systematically compare against different head geometry estimation and portrait manipulation approaches in the literature using 2D/3D warping and GANs, and demonstrate the superior performance of our method.
	\end{itemize}

	\section{Related Work}
	
	\paragraph{Face and hair 3D reconstruction.} 3D face reconstruction has been a longstanding task. Recently, deep 3D face reconstruction~\cite{tewari2017mofa,tran2017regressing,guo2018cnn,genova2018unsupervised} has attracted considerable attention. Our method follows the unsupervised learning schemes~\cite{tewari2017mofa,genova2018unsupervised} that train a network without ground-truth 3D data. For hair modeling, traditional methods perform orientation-map based optimization and sometimes require manual inputs \cite{chai2013dynamic} or a 3D hair exemplar repository \cite{chai2016autohair}. { Liang \etal~\cite{liang2018video} and Hu \etal~\cite{hu2017avatar} leverage hairstyle database for automatic hair reconstruction.} A deep 3D hair reconstruction method was proposed in \cite{zhou2018hairnet}, but the reconstructed hair strand model are not aligned with the input image thus cannot be used for our purpose.
	
	\paravspace
	\paragraph{CNN-based portrait editing and synthesis.} Face image editing and synthesis have attracted considerable attention in the vision and graphics community and have seen fast growth
	with the deep learning technique. Most existing CNN-based methods are devoted to editing appearance attributes such as skin color~\cite{choi2018stargan}, facial expression~\cite{choi2018stargan,pumarola2018ganimation,song2018geometry,portenier2018faceshop,geng2018warp}, makeup~\cite{chang2018pairedcyclegan,li2018beautygan},
	age~\cite{zhang2017age,choi2018stargan,wang2018face},
	and some other local appearance attributes~\cite{portenier2018faceshop,dekel2018sparse,shu2017neural}.
	Few methods worked on head pose manipulation. Perhaps the most relevant works among them are those synthesizing novel views (\eg, frontal) from an input face image~\cite{huang2017beyond,bao2018towards,yin2017towards}. However, the goals of these methods are not portrait editing and they do not handle hair and background. Some recent methods~\cite{shen2019interpreting,deng2020disentangled} shows that face rotation can be achieved by GAN latent space embedding and editing, however they still cannot well preserve hair structure. 
	
	\paravspace
	\paragraph{2D warping based facial animation.} Some approaches have been proposed to animate a face image with 2D warping~\cite{averbuch2017bringing,geng2018warp,wiles2018x2face}. Averbuch-Elor \etal~\cite{averbuch2017bringing} proposed to animate an image by the transferring the 2D facial deformations in a driving video using anchor points. A refinement process is applied to add fine-scale details and hallucinate missing regions. A similar pipeline is proposed by Geng \etal~\cite{geng2018warp}, which uses a GAN to refine the warped images. Wiles \etal~\cite{wiles2018x2face} proposes to generate 2D warping fields using neural networks.
	Lacking guidance from 3D geometry, there is no guarantee the face structure can be persevered by these 2D warping methods especially when head pose changes.
	
	\paravspace
	\paragraph{3D-guided view synthesis and facial animation.} 3D-guided face image frontalization and profiling have been used in different domains such as face recognition~\cite{taigman2014deepface,zhu2015high,hassner2015effective} and face alignment~\cite{zhu2016face}. These methods often only focus on facial region or handle hair and background naively. The most sophisticated face rotation method is perhaps due to Zhu \etal~\cite{zhu2016face}, which considers the geometry of the surrounding regions of a face. 
	However, their heuristically-determined region and depth do not suffice for realistic synthesis, and the rotated results oftentimes exhibit obvious inconsistency with the raw portraits. 
	Several works~\cite{nagano2018pagan,geng20193d} have been presented to synthesize facial expression leveraging 3D models, but they do not consider hair geometry and cannot manipulate head pose.
	
	\paravspace
	\paragraph{Video and RGBD based face reenactment.}  Several works have been presented for face reenactment with video or RGBD inputs~\cite{thies2016face2face,kim2018deep,thies2018headon}. Thies \etal~\cite{thies2016face2face} transfer facial expression in a source actor video to a target actor video with the aid of 3D face reconstruction.
	Kim \etal~\cite{kim2018deep} train a deep network on each given video to fit the portrait appearances therein. An RGBD reenactment system is presented by Thies \etal~\cite{thies2018headon}.
	
	\begin{figure*}[t!]
		\centering
		\includegraphics[width=0.99\textwidth]{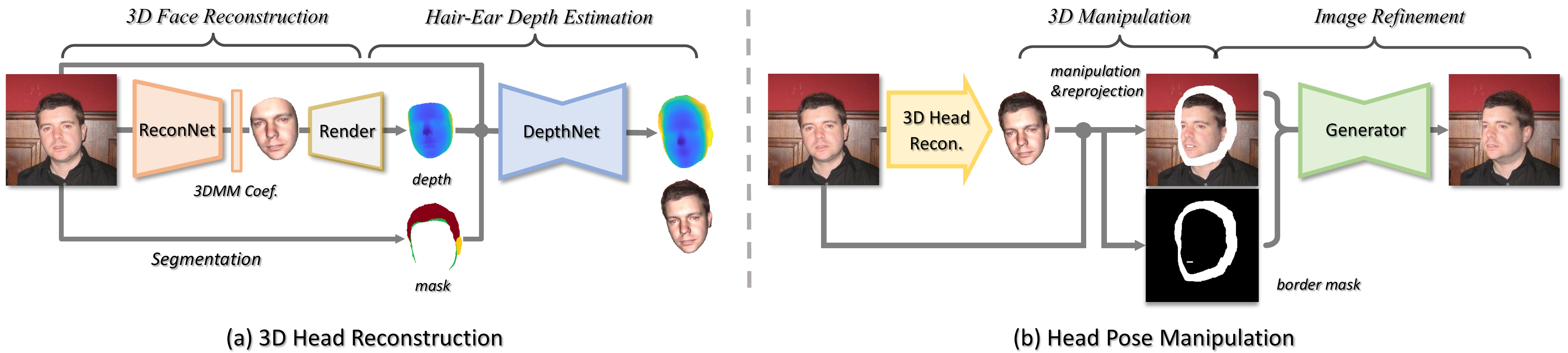}
		\vspace{-5pt}
		\caption{Overview of our single-image 3D head reconstruction and head pose manipulation methods.}\label{fig:framework}
		\vspace{-5pt}
	\end{figure*}
	
	\section{Overview and Preprocessing}
	
	The frameworks of our methods are depicted in Fig.~\ref{fig:framework}. After image preprocessing (to be described below), we run two-step 3D reconstruction with two CNNs to estimate 3D head pose and shape. For head pose manipulation, we first adjust the pose of the reconstructed shape in 3D and reproject it onto image plane, and then apply a refinement CNN to obtain the final result.
	\paravspace
	\paragraph{Preprocessing.} Given a portrait image, we perform rough alignment to centralize and rescale the detected face region (the image will be re-aligned later to accurately centralize the 3D head center after the 3D face reconstruction step). \textcolor[rgb]{0,0,0}{We then run a state-of-the-art face segmentation method of \cite{lin2019face} to segment out the head region, denoted as $\mathcal{S}$, which includes face, hair and ear regions.}

	\section{Single-Image 3D Head Reconstruction}
	
	In this work, we use the perspective camera model with an empirically-selected focal length. Head pose is determined by rotation $\bR \in \mathrm{SO}(3)$ and translation $\bt\in \mathbb{R}^3$ and is parameterized by $\bp\in\mathbb{R}^7$ with rotation represented by quaternion. We now present our method which reconstructs a 3DMM face as well as a depth map for other head regions.

	\subsection{Face Reconstruction and Pose Estimation}\label{sec:facerecon}
	
	With a 3DMM, the face shape $\bF$ and texture $\bT$ can be represented by an affine model:
	\begin{equation}
	\eqvspace
	\begin{split}
	\bF &= \bF(\balpha,\bbeta) = \bar{\bF} + \bB_{id}\balpha + \bB_{exp}\bbeta \\
	\bT &= \bT(\bdelta) = \bar{\bT} + \bB_t\bdelta
	\end{split}\label{equation:MM}
	\eqvspace
	\end{equation}
	where $\bar{\bF}$ and $\bar{\bT}$ are the average face shape and texture; $\bB_{id}$,  $\bB_{exp}$, and $\bB_t$ are the PCA bases of identity, expression, and texture respectively;
	$\balpha$, $\bbeta$, and $\bdelta$ are the corresponding coefficient vectors. We adopt the Basel Face Model~\cite{paysan20093d} for $\bar{\bF}$, $\bB_{id}$, $\bar{\bT}$, and $\bB_t$, and use the expression bases $\bB_{exp}$ of \cite{guo2018cnn} which are built from FaceWarehouse~\cite{cao2014facewarehouse}. After selection of basis subsets, we have $\balpha\in \mathbb{R}^{80}$, $\bbeta \in \mathbb{R}^{64}$ and $\bdelta \in \mathbb{R}^{80}$.
	
	Since ground-truth 3D face data are scarce, we follow recent methods~\cite{tewari2017mofa,genova2018unsupervised,deng2019accurate} to learn reconstruction in an unsupervised fashion using a large corpus of face images. 
	Our method is adapted from \cite{deng2019accurate} which uses hybrid-level supervision for training. 
	Concretely, the unknowns to be predicted can be represented by a vector $(\balpha,\bbeta,\bdelta,\bp,\bgamma)\in\mathbb{R}^{239}$, where $\bgamma\in\mathbb{R}^{9}$ is the Spherical Harmonics coefficient vector for scene illumination.
	Let $I$ be a training image and $I'$ its reconstructed counterpart rendered with the network prediction, we minimize the photometric error via:
	\vspace{-3pt}
	\begin{equation}
	\vspace{-6pt}
	\eqvspace
	l_{photo} = \text{\scriptsize $\int\!\!\!$}_{\mathcal{F}}\|I-I'(\balpha,\bbeta,\bdelta,\bgamma,\bp)\|_2 \label{eq:face_photo}
	\eqvspace
	\vspace{0pt}
	\end{equation}
	where $\mathcal{F}$ denotes the rendered face region we consider here\footnote{\textcolor[rgb]{0,0,0}{For brevity, in our loss functions we drop the notation of pixel variable in the area integral. We also drop the normalization factors (\eg, $\frac{1}{N_\mathcal{F}}$ in Eq.~\ref{eq:face_photo} where $N_\mathcal{F}$ is the number of pixels in region $\mathcal{F}$).} }, and $\|\cdot\|_2$ denotes the $\ell_2$ norm for residuals on r, g, b channels.
	We also minimize the perceptual discrepancy between the rendered and real faces via:
	\vspace{-3pt}
	\begin{equation}
	\eqvspace
	l_{per} = 1-\frac{<f(I),f(I')>}{\|f(I)\|\cdot\|f(I')\|}
	\eqvspace
	\vspace{-6pt}
	\end{equation} where $f(\cdot)$ denotes a face recognition network for identity feature extraction where the model from \cite{yang2017neural} is used here. Other commonly-used losses such as the 2D facial landmark loss and coefficient regularization loss are also applied, and we refer the readers to \cite{deng2019accurate} for more details.
	
	\subsection{Hair\&Ear Depth Estimation}
	
	Our next step is to estimate a depth map for other head region, defined as $\mathcal{H}=\mathcal{S}-(\mathcal{S}^f\bigcap\mathcal{F})$ where {$\mathcal{S}^f$ denotes the face region defined by segmentation.} $\mathcal{H}$ includes hair and ear as well as a small portion of segmented face region that is not covered by the projected 3DMM face.
	Due to lack of ground-truth depth data, we train a network using a collection of image pairs in a stereo matching setup. Note we use image pairs only for training purpose. The network always runs on a single image at test time.
	
	Let $I_1$, $I_2$ be a training image pair of one subject (\eg, two frames from a video) with different head poses $(\bR_1, \bt_1)$, $(\bR_2, \bt_2)$ recovered by our face reconstruction network. Our goal is to train a single network to predict both of their depth maps $\depthmap_1$ and $\depthmap_2$ in a siamese network scheme~\cite{chopra2005learning}. Before training, we first run naive triangulation on regular pixel grids of $\mathcal{H}_1$ and $\mathcal{H}_2$ to build two 2D meshes. Given depth map estimate $\depthmap_1$, a 3D mesh $\mathbf{H}_1$ can be constructed via inverse-projection. We can transform $\mathbf{H}_1$ to $I_2$'s camera system via $(\bR_2\bR_1^{-1}, -\bR_2\bR_1^{-1}\bt_1+\bt_2)$, and project it onto image plane to get a synthesized image $I_2'$. Similar process can be done for generating $I_1'$ from $I_2$ and $\depthmap_2$. The whole process is differentiable and we use it to train our depth prediction network with the following losses.
	
	As in stereo matching, we first enforce color constancy constraint by minimizing the brightness error
	\begin{equation}
	\eqvspace
	l_{color} = \text{\scriptsize $\int\!\!\!$}_{\mathcal{H}_2'}\|I_2'(\depthmap_1) - I_2\|_1 + \text{\scriptsize $\int\!\!\!$}_{\mathcal{H}_1'}\|I_1'(\depthmap_2) - I_1\|_1
	\eqvspace
	\end{equation}
	where $\mathcal{H}_2'=\mathcal{H}_2'(\mathcal{H}_1, \depthmap_1)$ is the warped region from $\mathcal{H}_1$ computed by head poses and $\depthmap_1$ in the transformation process described above; similarly for $\mathcal{H}_1'=\mathcal{H}_1'(\mathcal{H}_2, \depthmap_2)$.
	We also apply a gradient discrepancy loss which is robust to illumination change thus widely adopted in stereo and optical flow estimation~\cite{brox2004high,bleyer2011patchmatch,yang2015dense}:
	\begin{equation}
	\eqvspace
	\!\!\!l_{grad}\!=\!\!\text{\scriptsize $\int\!\!\!$}_{\mathcal{H}_2'}\!\|\nabla I_2'(\depthmap_1) -\! \nabla I_2\|_1 \!+\!\text{\scriptsize $\int\!\!\!$}_{\mathcal{H}_1'}\!\|\nabla I_1'(\depthmap_2) -\! \nabla I_1\|_1\!\!\!
	\eqvspace
	\end{equation}
	where $\nabla$ denotes the gradient operator. To impose a spatial smoothness prior, we add a second-order smoothness loss
	\begin{equation}
	\eqvspace
	l_{smooth} = \text{\scriptsize $\int\!\!\!$}_{\mathcal{H}_1}|\Delta \depthmap_1| + \text{\scriptsize $\int\!\!\!$}_{\mathcal{H}_2}|\Delta \depthmap_2|
	\eqvspace
	\end{equation}
	where $\Delta$ denotes the Laplace operator. 
	\paravspace
	\paragraph{Face depth as condition and output.} Instead of directly estimating hair and ear depth from the input image $I$, we project the reconstructed face shape $\bF$ onto image plane to get a face depth map $\depthmap^f$. We make $\depthmap^f$ an extra conditional input concatenated with $I$. Note $\depthmap^f$ provides beneficial information (\eg, head pose, camera distance) for hair and ear depth estimation. In addition, it allows the known face depth around the contour to be easily propagated to the adjacent regions with unknown depth.
	
	More importantly, we train the network to also predict the depth of the facial region using $\depthmap^f$ as target:
	{
		\begin{equation}
		\eqvspace
		l_{face} = \!=\!\text{\scriptsize $\int\!\!\!$}_{\mathcal{F}_1 - \mathcal{S}^h_1\bigcap\mathcal{F}_1 }\!|\depthmap_1 - \depthmap^f_1| \!+\! \text{\scriptsize $\int\!\!\!$}_{\mathcal{F}_2 - \mathcal{S}^h_2\bigcap\mathcal{F}_2 }\!|\depthmap_2 - \depthmap^f_2|\!
		\eqvspace
		\end{equation}
		where $\mathcal{S}^h$ denotes the hair region defined by segmentation.} Note learning face depth via $l_{face}$ should not introduce much extra burden for the network since $\depthmap^f$ is provided as input. But crucially, we can now easily enforce the consistency between the reconstructed 3D face and the estimated 3D geometry in other regions, as in this case we calculate the smoothness loss across whole head regions $\mathcal{S}_1$, $\mathcal{S}_2$:
	\begin{equation}
	\eqvspace
	l_{smooth} = \text{\scriptsize $\int\!\!\!$}_{\mathcal{S}_1}|\Delta \depthmap_1| + \text{\scriptsize $\int\!\!\!$}_{\mathcal{S}_2}|\Delta \depthmap_2|
	\eqvspace
	\end{equation}
	
\begin{figure}[t!]
		\small
		\centering
		\includegraphics[width=\columnwidth]{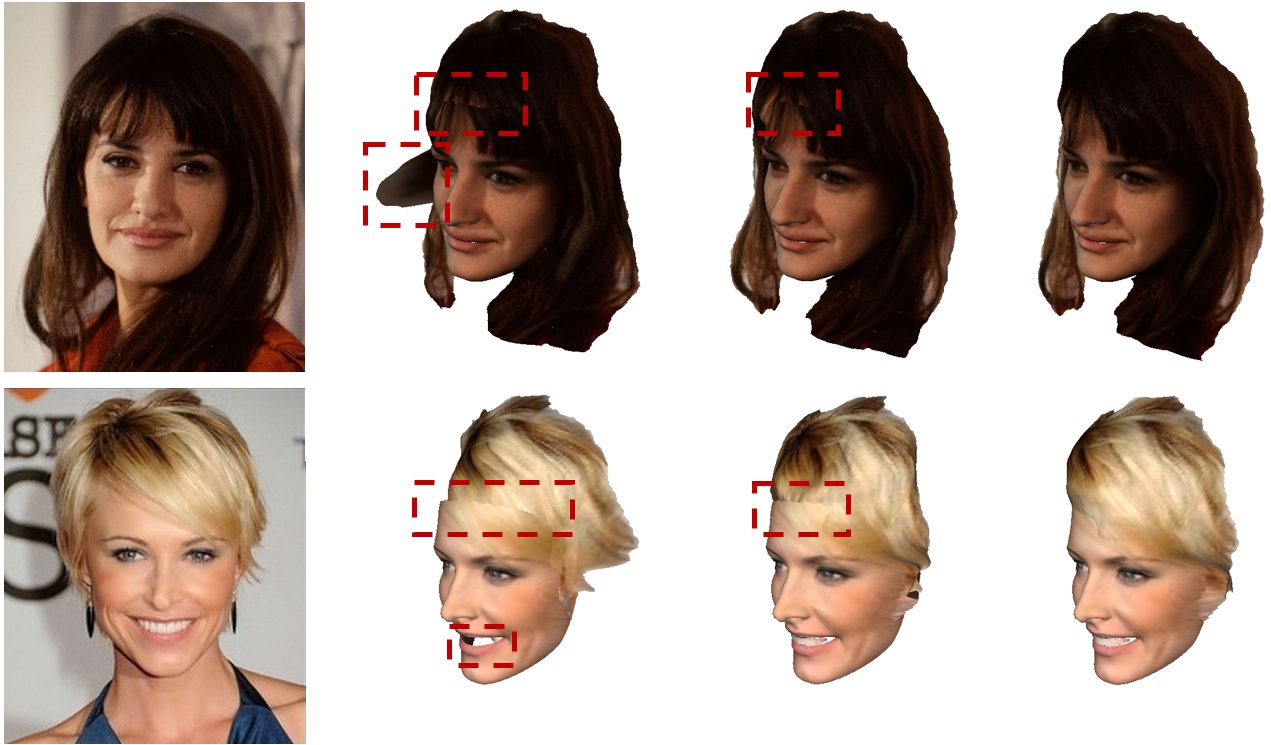}\\
		\ \ \ \ Input\ \ \ \ \ \ \ \ \ \ \ \ \  w/o face depth \ \  \ with face depth\ \ \ \ \ \ \ + $l_{layer}$ \\
		\vspace{-5pt}
		\caption{\textcolor[rgb]{0,0,0}{3D head reconstruction result of our method with different settings.}}\label{fig:depthestimation_ablation}
		\vspace{-2pt}
	\end{figure}

	Figure~\ref{fig:depthestimation_ablation} (2nd and 3rd columns) compares the results with and without face depth. We also show quantitative comparisons in Table~\ref{table:eval} (2nd and 3rd columns). As can be observed, using face depth significantly improves head geometry consistency and reconstruction accuracy.
	
	\paravspace
	\paragraph{Layer-order loss.}
	{Hair can often occlude a part of facial region, leading to two depth layers. To ensure correct relative position between the hair and occluded face region (\ie, the former should be in front of the latter), we introduced a layer-order loss defined as:
		\begin{equation}
		\eqvspace
		l_{layer}\! = \!\text{\scriptsize$\int\!\!\!$}_{\mathcal{S}^h_1\bigcap\mathcal{F}_1}\!\!\!\max(0, \depthmap_1 - \depthmap^f_1)\!+\!\text{\scriptsize$\int\!\!\!$}_{\mathcal{S}^h_2\bigcap\mathcal{F}_2}\!\!\!\max(0, \depthmap_2 - \depthmap^f_2)
		\eqvspace
		\end{equation}which penalizes incorrect layer order.
		As shown in Fig.~\ref{fig:depthestimation_ablation}, the reconstructed shapes are more accurate with $l_{layer}$.
	}
	
	\paravspace
	\paragraph{Network structure.} We apply a simple encoder-decoder structure using a ResNet-18~\cite{he2016deep} as backbone. We discard its global average pooling and the last fc layers, and append several transposed convolutional layers to upsample the feature maps to the full resolution. Skip connections are added at $64\times 64$, $32\times 32$ and $16\times 16$ resolutions. The input image size is $256\times 256$. More details of the network structure can be found in the \emph{suppl. material}.

	\section{Single-Image Head Pose Manipulation}
	Given the 3D head model reconstructed from the input portrait image, we modify its pose and synthesize new portrait images, described as follows.
	
	\subsection{3D Pose Manipulation and Projection}
	To change the head pose, one simply needs to apply a rigid transformation in 3D for the 3DMM face $\bF$ and hair-ear mesh $\bH$ given the target pose $\bar{\bp}$ or displacement $\delta\bp$. 
	After the pose is changed, we reproject the 3D model onto 2D image plane to get coarse synthesis results. \textcolor[rgb]{0,0,0}{Two examples are shown in Fig.~\ref{fig:gan_ablation}}.
	
	\subsection{Image Refinement with Adversarial Learning}
	
	The reprojected images suffer from several issues. Notably, due to pose and expression change, some holes may appear, where the missing background and/or head region should be hallucinated akin to an image inpainting process.
	Besides, the reprojection procedure may also induce certain artifacts due to imperfect rendering.
	
	\begin{figure}[t!]
		\small
		\centering
		\includegraphics[width=\columnwidth]{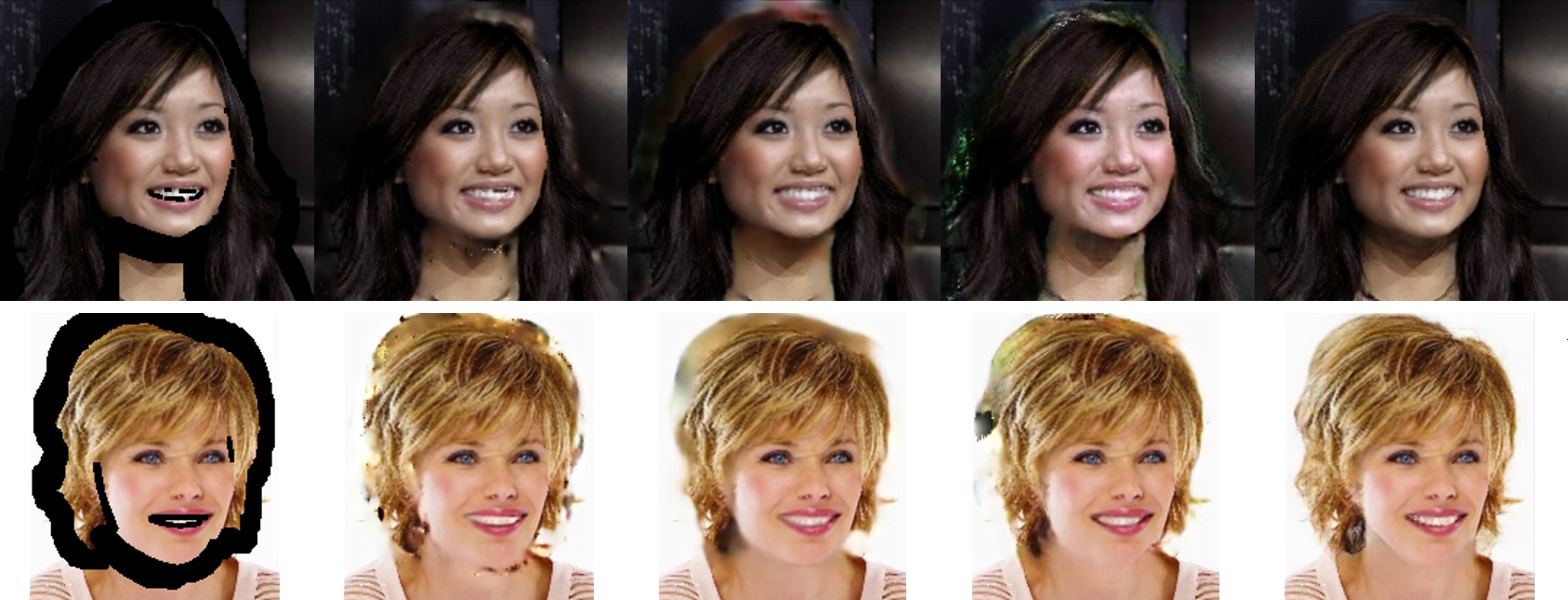}\\
		Input \ \ \ \ \ Mixing output \ \ \ \ \ $l_{color}$ \ \ \ \ \ \ $l_{color}\!+\!l_{adv}$ \ \ \ \ \ Ours \ \ \ \\
		\vspace{-5pt}
		\caption{Results of the image refinement network trained with different settings.}\label{fig:gan_ablation}
		\vspace{-5pt}
	\end{figure}
	
	\begin{figure*}[t!]
		\vspace{-3pt}
		\centering
		\includegraphics[width=0.985\linewidth]{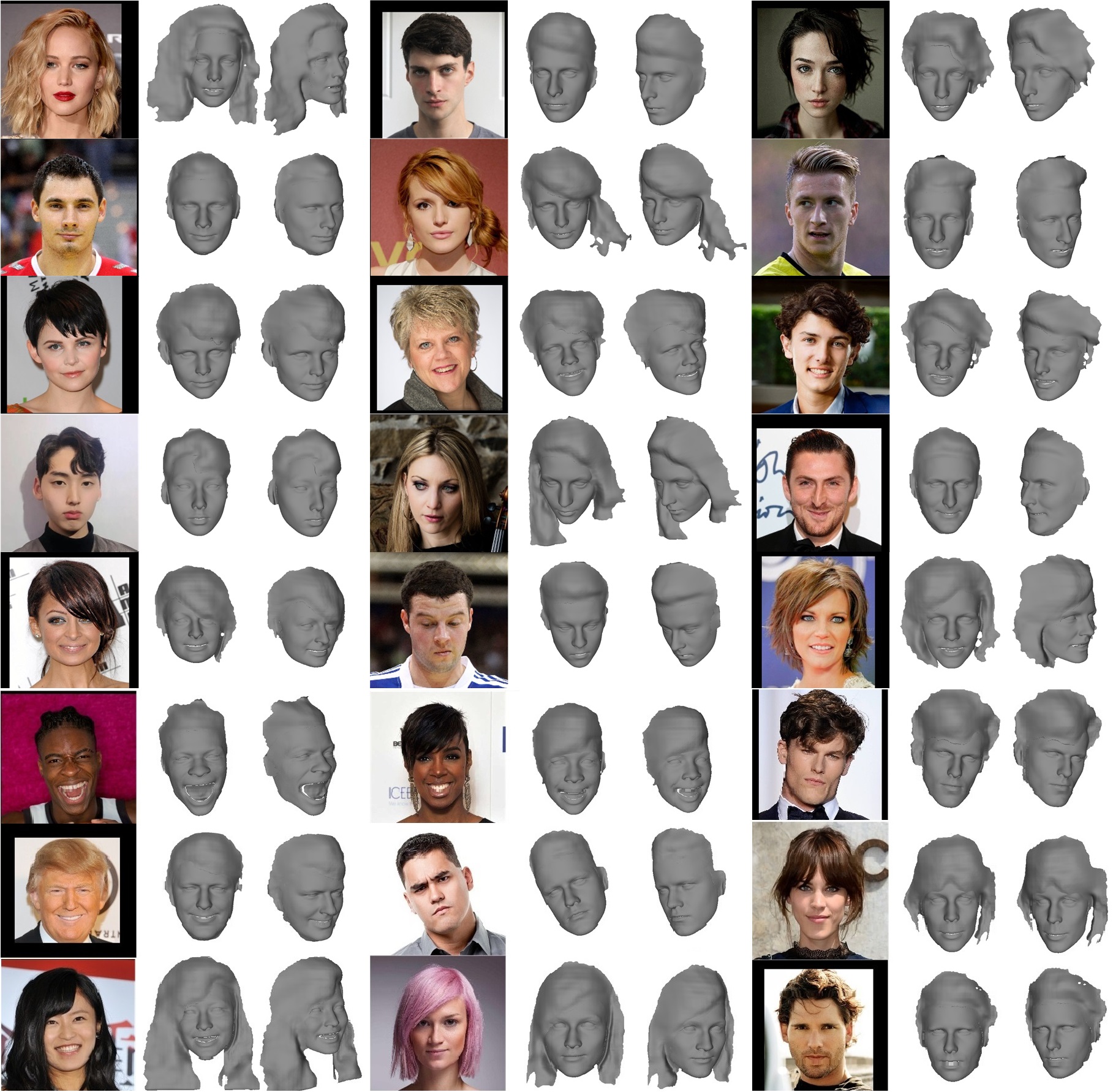}
		\vspace{-6pt}
		\caption{Typical single-image head reconstruction results. Our method can deal with a large variety of  face shapes and hair styles, generating high-quality 3D head models. Note our method is trained without any ground-truth 3D data. 
		}\label{fig:results_3d}
		\vspace{-4pt}
	\end{figure*}
	\begin{figure*}[t!]
		\centering
		\includegraphics[width=0.99\textwidth]{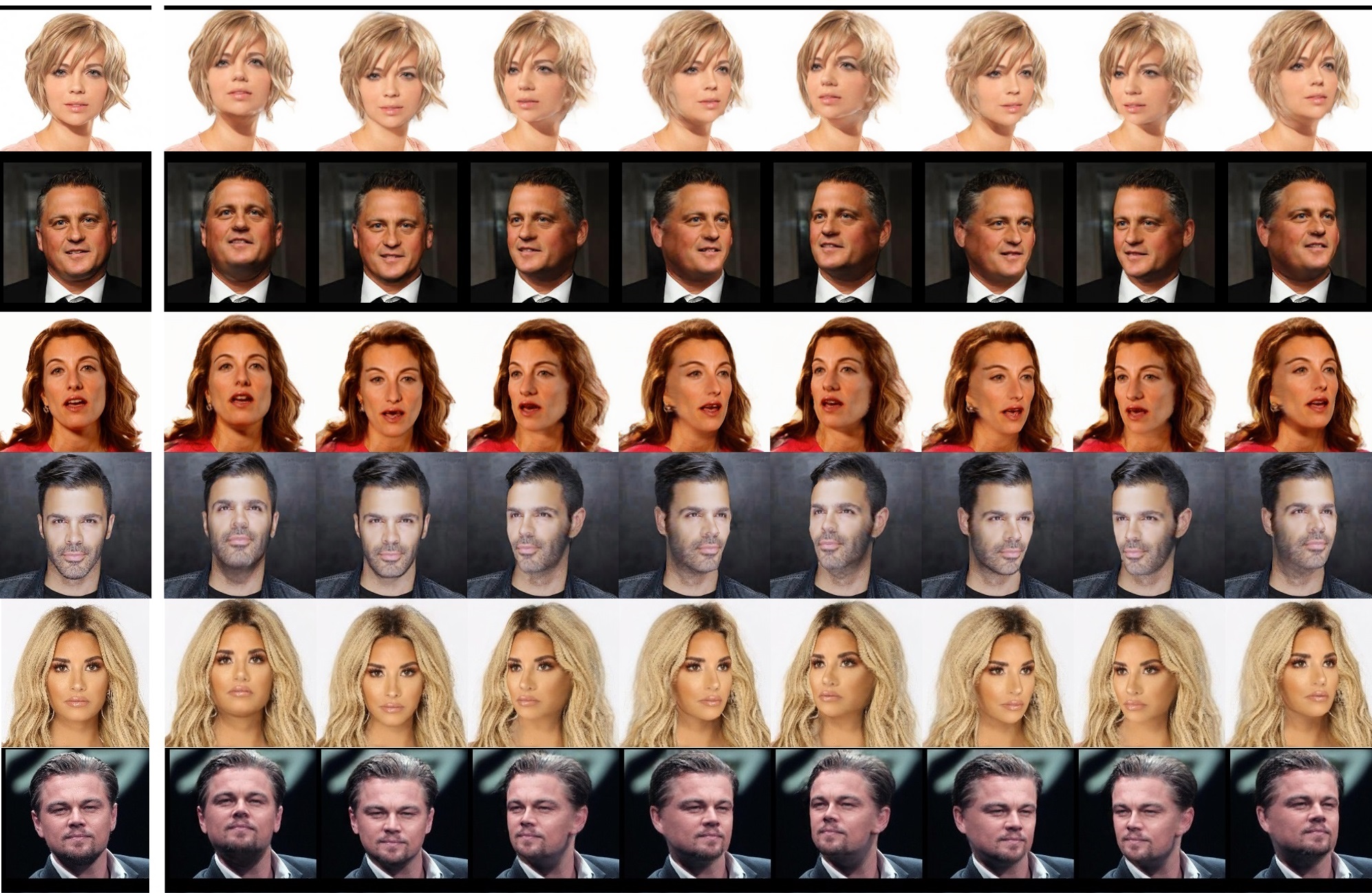}
		\vspace{-7pt}
		\caption{Typical pose manipulation results. The left column shows the input images to our method, and the other columns show our synthesized images with altered head poses.\label{fig:poses}}
		\vspace{-9pt}
	\end{figure*}
	To address these issues, we apply a deep network $G$ to process these images. For stronger supervision, we leverage both paired (\ie, images with ground truth label) and unpaired data (\ie, our coarse results) to train such a network. To obtain paired data, we take some real images with various head poses, and synthetically masked out some regions along the head segmentation boundaries. Let $J$ be an unpaired coarse result, and ($R, \hat{R}$) be the paired data where $R$ denotes the corrupted image and $\hat{R}$ its corresponding real image, we apply the $\ell_1$ color loss via
	{\small
		\begin{equation}
		\eqvspace
		\!\!\!l_{color}(G)\!=\!\mathbb{E}_J\big[\text{\scriptsize $\int\!\!\!$}_\mathcal{B} \|G(J)-J\|_1\big] + \E_R \big[\text{\scriptsize $\int$}\|G(R)-\hat{R}\|_1\big]
		\eqvspace
		\end{equation}
	}where $\mathcal{B}$ denotes the background and the warped head regions of $J$. 
	
	We apply adversarial learning to improve the realism of the generated images. We introduce a discriminator $D$ to distinguish the outputs of $G$ from real images, and train $G$ to fool $D$. The LS-GAN~\cite{mao2017least} framework is used, and our adversarial loss functions for $G$ and $D$ can be written as 
	{\small
		\begin{eqnarray}
		\eqvspace
		\small
		l_{adv}(G)\!=\!\frac{1}{2}\mathbb{E}_J\!\big[(D(G(J))\!-\!1)^2\big]\!+\!\frac{1}{2}\mathbb{E}_R\!\big[(D(G(R))\!-\!1)^2\big],\!\!
		\eqvspace
		\end{eqnarray}
		\vspace{-5pt}
	}
	{\small
		\begin{equation}
		\eqvspace
		\begin{aligned}
		\!\!\!\!\!l_{adv}(D) \!= & \frac{1}{2}\E_J\!\!\big[(D(G(J))\!-\!0)^2\big]\!+\!\frac{1}{2}\!\E_R\!\big[(D(G(R)) \!-\! 0)^2\big]\!\! \\
		& +\!\E_R\!\big[(D(\hat{R})\!-\!1)^2\big],
		\end{aligned}
		\eqvspace
		\end{equation}
	}respectively. As shown in Fig.~\ref{fig:gan_ablation}, with the aid of the adversarial loss, our model generates much sharper results. However, some unwanted artifacts are introduced, possibly due to unstable GAN training.
	
	To remove these artifacts, we further apply a deep feature loss, also known as perceptual loss~\cite{johnson2016perceptual}, for paired data via
	{\small
		\begin{equation}
		\eqvspace
		\!\!\!l_{feat}(G)\!=\!\sum_{i}\frac{1}{N_{i}}\|\phi_{i}(G(R))-\phi_{i}(\hat{R})\|_1
		\eqvspace
		\end{equation}
	}where $\phi_{i}$ is the $i$-th activation layer of the VGG-19 network~\cite{simonyan2014very} pretrained on ImageNet~\cite{deng2009imagenet}.  We use the first layers in all blocks.  Figure~\ref{fig:gan_ablation} shows that our final results appear quite realistic. They are sharp and artifact-free.
	
	\paravspace
	\paragraph{Difference with image inpainting.}
	In our task, the reprojected head portrait, though visually quite realistic to human observer, may contain some unique characteristics that can be detected by the discriminator. We tried generating refined results to be fed into $D$ by mixing $G$'s partial output and the raw input -- a popular formulation in deep image inpainting~\cite{yu2018generative} -- via $J'=M\odot G(J)+(1-M)\odot J$ where $M$ is the missing region mask. However, the results are consistently worse than our full image output strategy (see Fig.~\ref{fig:gan_ablation} for a comparison).
	
	\paravspace
	\paragraph{Network structure.}
	Our network structures for $G$ and $D$ are adapted from \cite{wang2018high}. 
	The input and output image size is $256\times 256$. More details can be found in the \emph{suppl. material}.
	
	\section{Experiments}
	\paragraph{Implementation Details.~}
	Our method is implemented with Tensorflow~\cite{tensorflow}.\footnote{Code and trained model will be released publicly.} 
	The face reconstruction network is trained with 180K in-the-wild face images from multiple sources such as CelebA~\cite{liu2015deep}, 300W-LP~\cite{zhu2016face} and LFW~\cite{huang2007labeled}.
	To train the head depth network, we collected 11K image pairs from 316 videos of 316 subjects that contain human head movements\footnote{We assume hair undergoes rigid motion within small time windows.}. The relative rotations are mostly within 5 to 15 degrees. The training took 15 hours on 1 NVIDIA M40 GPU card.
	To train the image refinement network, we collected 37K paired data and 30K unpaired data, and the training took 40 hours on 4 M40 GPU cards. Due to space limitation, more implementation details and results are shown in the \emph{suppl. material}.

	\subsection{Results}
	
	The results from our method will be presented as follows. Note all the results here are from our test set where the images were not used for training.
	
	\paravspace
	\paragraph{3D head reconstruction.}
	
	Figure~\ref{fig:results_3d} shows some typical samples of our single-image 3D head reconstruction results. As can be observed, our reconstruction networks can produce quality face and hair geometry given a single portrait image, despite we did not use any ground-truth 3D data for training. Various hair styles can be well handled as shown in the figure. Although the face region and the hair-ear part have different, disjoint representations (3DMM \emph{vs.} depth map) and are reconstructed in two steps, they appear highly consistent with each other and the resultant head models are visually pleasing. 
	
	\begin{table}[t!]
		\centering
		\caption{\label{table:eval}Average 3D reconstruction error evaluated on RGBD images from the Biwi dataset~\cite{fanelli_IJCV}.}
		{\small
			\vspace{-6pt}
			\begin{tabular*}{0.68\columnwidth}{lccc}
				\hline
				Error (mm)& \!\!\!Zhu \cite{zhu2016face}\!\! & \!\!Ours$_{w\!/\!o~d^{f}}$\!\! & \!Ours\! \\ \hline
				Face         & 5.05 & 4.31 & \textbf{3.88} \\ \hline
				Non-face     & 8.56 & 7.39  & \textbf{6.78}\\ \hline
			\end{tabular*}
			\vspace{-5pt}
		}
	\end{table}
	
	{
		For quantitative evaluation and ablation study, we use the RGBD images from Biwi Kinect Head Pose Database~\cite{fanelli_IJCV} which contains 20 subjects with various hair styles. 
		We compute the head reconstruction errors of our method using the depth images as ground-truth geometry. The error is computed as the average point distances in 3D between the outputs and ground-truth shapes after 3D alignment. The results are presented in Table~\ref{table:eval}, which shows the decent 3D reconstruction accuracy of our method. It also shows that the accuracy decreases if  face depth is not used as the input for the depth estimation network, demonstrating the efficacy of our algorithm design.
	}
	
	\paravspace
	\paragraph{Pose Manipulation.}
	
	Figure~\ref{fig:poses} presents some pose manipulation results from our method. It can be seen that our method can generate realistic images with new head poses.
	Not only the face identity is well preserved, but also the hair shapes are highly consistent across different poses. The background is not disrupted by pose change.

	\subsection{Comparison with Prior Art}\label{sec:compare_previous}
	
	\begin{figure}[t!]
		\small
		\centering
		\includegraphics[width=0.99\columnwidth]{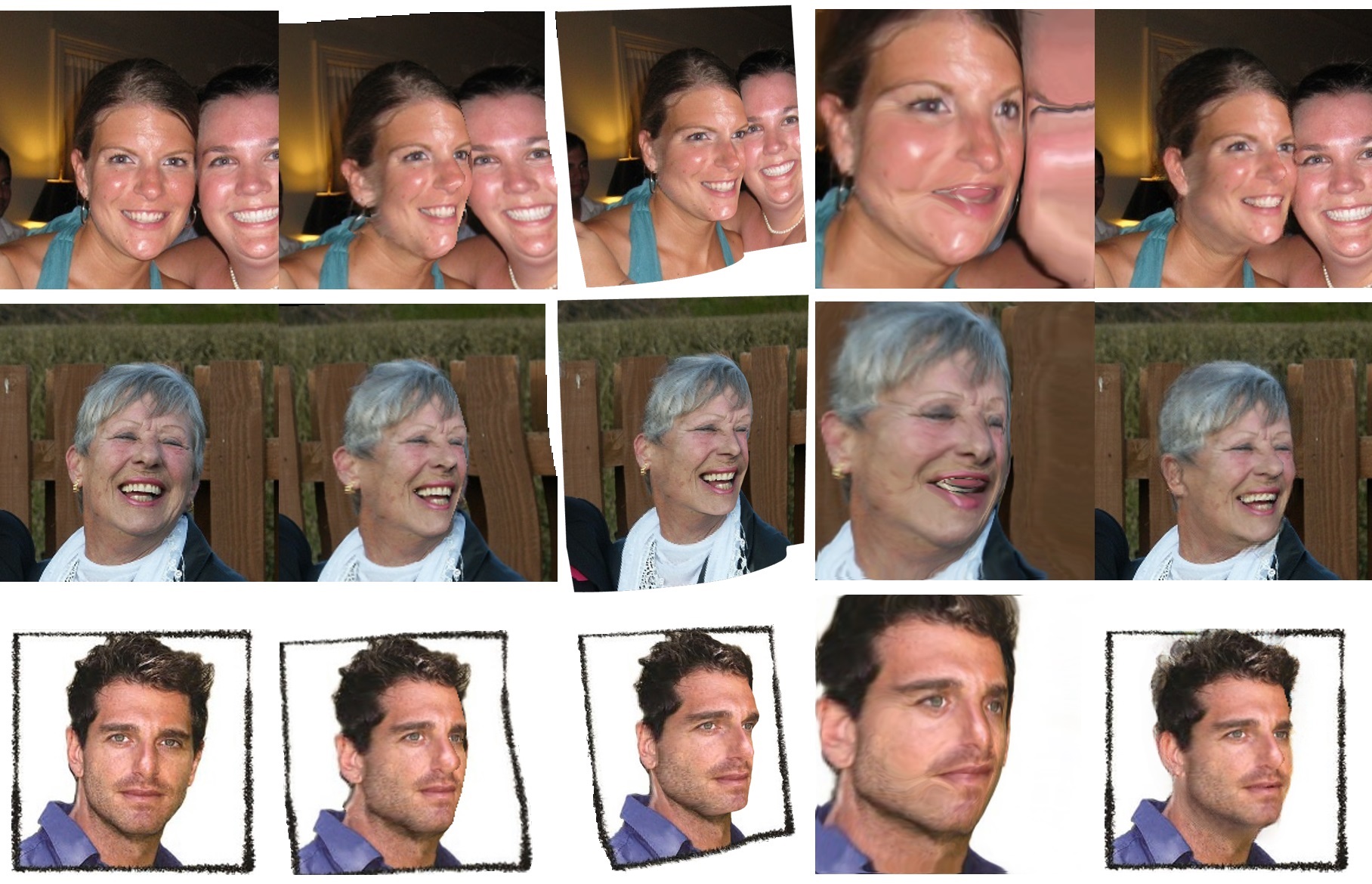}\vspace{-2pt}\\
		\ Input\ \ \ \ \ \ \ \ \ \ \ \ \ \ \cite{zhu2016face} \ \ \ \ \ \ \ \ \ \ \ \ \ \ \ \cite{chai2015high} \ \ \ \ \ \ \ \ \  \ \ \ \ \cite{wiles2018x2face} \ \ \ \ \ \ \ \ \ \ \ \ Ours\\
		\vspace{-6pt}
		\caption{Comparison with the methods of Zhu \etal~\cite{zhu2016face}, Chai \etal~\cite{chai2015high}, and Wiles \etal~\cite{wiles2018x2face}.
			\label{fig:compare_3dwarping}}
		\vspace{-2pt}
	\end{figure}
	
	\paragraph{Comparison with Zhu \etal~\cite{zhu2016face}.}
	Zhu \etal~\cite{zhu2016face} proposed a CNN-based 3D face reconstruction and alignment approach for single images. It also provides a warping-based portrait rotation method, originally developed for training data generation, based on 3D face geometry. To obtain reasonable warping for hair and ear regions, it defines a surrounding region of the face and heuristically determines its depth based on face depth. Figure~\ref{fig:compare_3dwarping} compares the face rotation results of \cite{zhu2016face} and ours. It can be seen that \cite{zhu2016face}'s results may suffer from obvious distortions. In contrast, our method can generate new views that are not only more realistic but also more consistent with the input images.
	
	Also note that \cite{zhu2016face} simply warps the whole image including the background region. Background change is undesired for portrait manipulation but is commonly seen in previous 2D/3D warping based methods~\cite{chai2015high,zhu2015high,zhu2016face,averbuch2017bringing}. In contrast, our method can well preserve the background.
	
	Table~\ref{table:eval} compares the 3D reconstruction error of \cite{zhu2016face} and our method using images from the Biwi dataset~\cite{fanelli_IJCV}. It shows that our method outperforms \cite{zhu2016face} by an appreciable margin: our errors are 23.17\% and 20.79\% lower in face and non-face regions, respectively.
	
	\begin{table}[t!]
		\centering
		\caption{\label{tab:featuresim}Average perceptual similarity (deep feature cosine similarity) between the input and rotated face images.}
		\vspace{-6pt}
		{\small
			
			\begin{tabular*}{0.7\columnwidth}{ccc}
				\hline
				& Chai \etal~\cite{chai2015high} & Ours  \\
				\hline
				Cosine distance    & 0.829      & \textbf{0.856}  \\
				\hline
			\end{tabular*}
			\vspace{-2pt}
		}
	\end{table}
	
	\paravspace
	\paragraph{Comparison with Chai \etal~\cite{chai2015high}.}
	We then compare with Chai \etal~\cite{chai2015high}, which is a traditional optimization-based method developed for hair modeling. It also estimates face shape by facial landmark fitting.
	We run the program released by \cite{chai2015high}, which requires a few user-provided strokes before reconstruction and provides 3D view of the input image after running reconstruction. 
	As shown in Fig.~\ref{fig:compare_3dwarping}, the method of \cite{chai2015high} also leads to some distortions, while our results are more consistent with the input faces. The differences in background regions are also prominent.
	
	For quantitative comparison, we consider a face recognition setup and compute the perceptual similarity between the original images and the warped ones. For fair comparison, we use the 10 images shown in \cite{chai2015high} (Figure 13) and the corresponding results therein. For our method, we rotate the raw faces to same poses as theirs. We use VGG-Face~\cite{parkhi2015deep} for deep face feature extraction, and Table~\ref{tab:featuresim} shows the higher perceptual similarity of our results.
	
	\begin{figure}[t!]
		\vspace{0pt}
		\centering
		\small
		\begin{center}
			\includegraphics[width=1\linewidth]{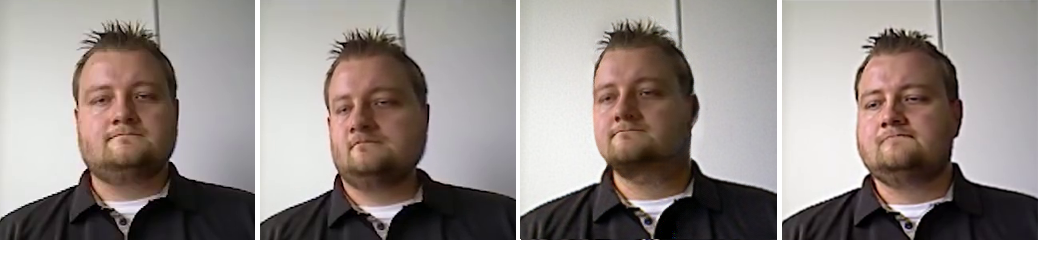}\vspace{-2pt}\\
			Input~~~~~~~~~~~~~~~~~~~\cite{averbuch2017bringing}~~~~~~~~~~~~~~~~~~~~~Ours~~~~~~~~~~~~~~~~~~~\cite{thies2018headon}
		\end{center}
		\vspace{-17pt}
		\caption{Comparison with Averbuch-Elor \etal~\cite{averbuch2017bringing}. The input image and result of \cite{averbuch2017bringing} are from \cite{thies2018headon}.}
		\label{fig:headon}
		\vspace{3pt}
	\end{figure}
	
	\paravspace
	\vspace{-2pt}
	\paragraph{Comparison with Averbuch-Elor~\etal~\cite{averbuch2017bringing}.}
	In Fig.~\ref{fig:headon}, we qualitatively compare with a 2D-warping based face reenactment method of Averbuch-Elor~\etal~\cite{averbuch2017bringing}, which drives a face image with a reference face video for animation.
	As can be seen, pose change is problematic for 2D warping and the result exhibits obvious distortions. Ours contains much less distortion and appears more realistic. For reference, we also present in Fig.~\ref{fig:headon} the result of Thies \etal~\cite{thies2018headon}, which works on RGBD images and builds a target actor 3D model offline for high-quality reenactment.
	
	\begin{figure}[t!]
		\vspace{-3pt}
		\small
		\centering
		\includegraphics[width=0.99\columnwidth]{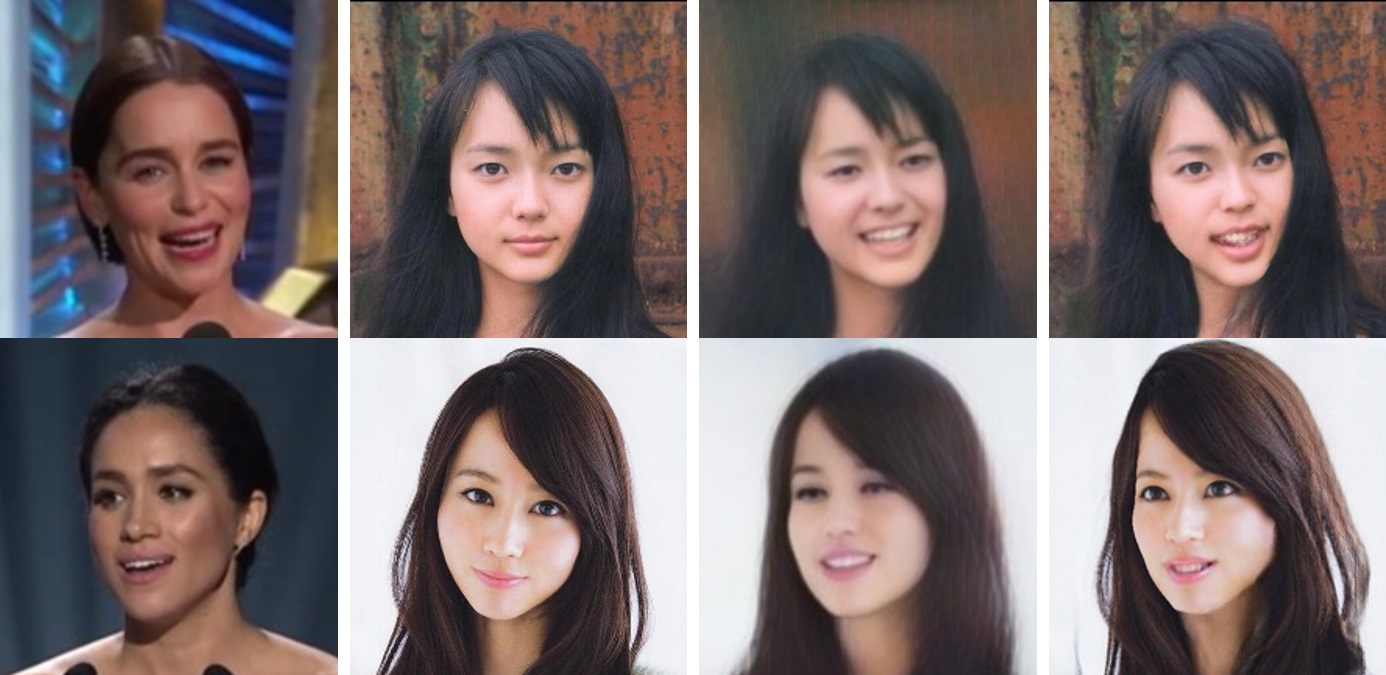}
		~~~Source~~~~~~~~~~~~~~~Target~~~~~~~~~~~FSGAN~\cite{nirkin2019fsgan}~~~~~~~~~~~~Ours
		\vspace{-7pt}
		\caption{Comparison with the FSGAN method of Nirkin~\etal~\cite{nirkin2019fsgan}. Images are from~\cite{nirkin2019fsgan}.
		}\label{fig:fsgan_x2face}
		\vspace{-10pt}
	\end{figure}
	
	\paravspace
	\vspace{-2pt}
	\paragraph{Comparison with X2Face~\cite{wiles2018x2face}.} We also compare with another 2D warping-based face manipulation method, X2Face~\cite{wiles2018x2face}, where the warping fields are generated by neural networks. As shown in Fig.~\ref{fig:compare_3dwarping}, the results of \cite{wiles2018x2face} suffer from obvious distortions and cannot handle missing regions, whereas ours appear much more natural.
	
	{
		\paravspace
		\paragraph{Comparison with FSGAN~\cite{nirkin2019fsgan}.} Finally, we compare with a recent GAN-based face swapping and reenactment method, FSGAN~\cite{nirkin2019fsgan}. As shown in Fig.~\ref{fig:fsgan_x2face}, the results of \cite{nirkin2019fsgan} tend to be over-smoothed. We believe there's a still great hurdle for GANs to directly generate fine details that are geometrically correct, given the complex geometry of hairs. However, the expression of \cite{nirkin2019fsgan}'s results appears more vivid than ours especially for the first example. One of our future works would be integrating fine-grained expression manipulation into our pipeline.
		
	}
	
\begin{figure}[t!]
		
		\centering
		\includegraphics[width=0.85\columnwidth]{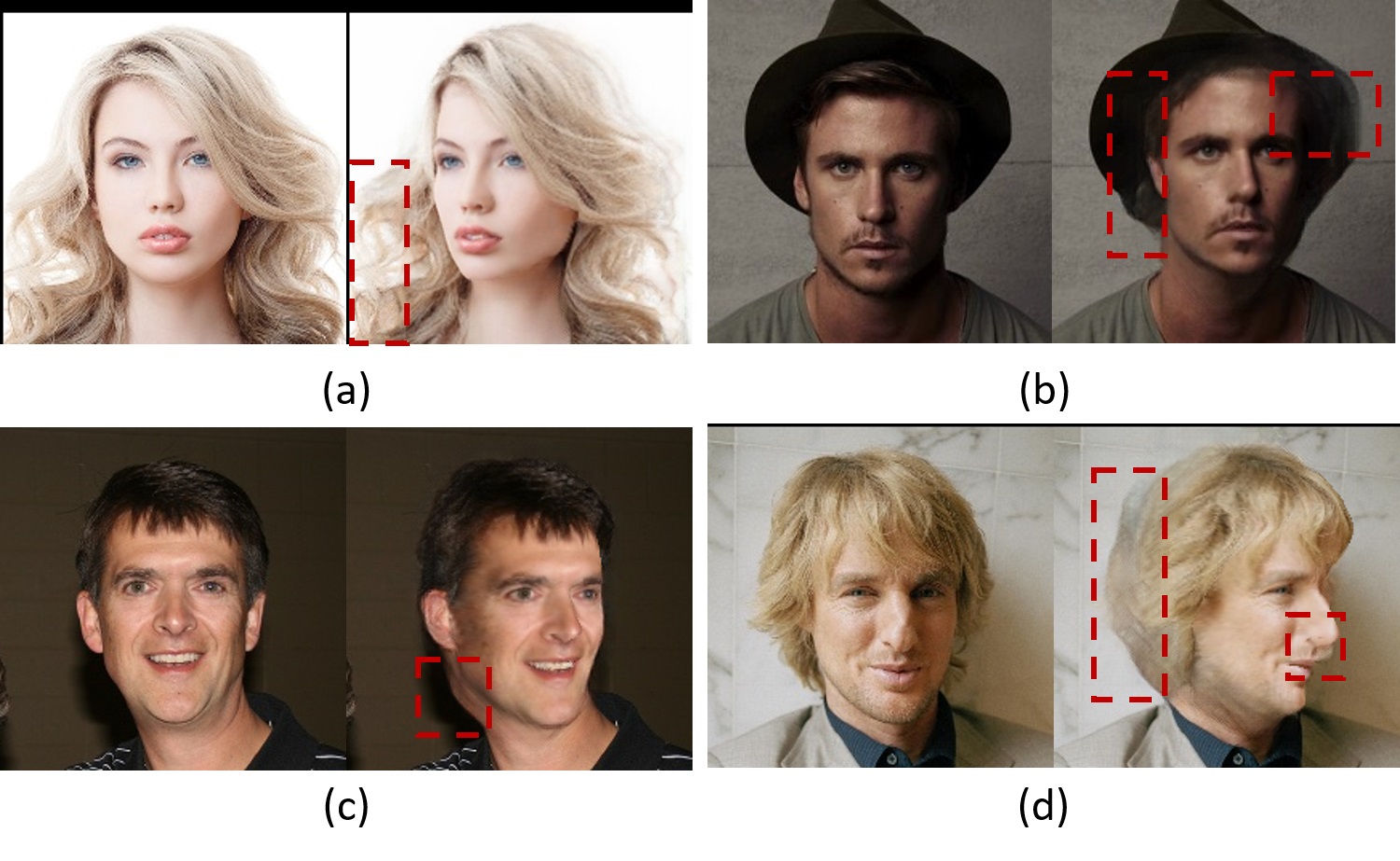}
		\vspace{-10pt}
		\caption{Failure cases due to (a) wrong segmentation, (b) obstruction, (c) inaccurate rotation center estimate, and (d) extreme pose.\label{fig:failurecases}}
	\end{figure}

	\paravspace
	\paragraph{Failure cases.}
	Our method may fail under several situations, as illustrated in Fig.~\ref{fig:failurecases}. For example, erroneous segmentation and obstructions may lead to apparent artifacts.
	Our head center is estimated in the face reconstruction step, and artifacts may appear for a few cases with inaccurate head center estimates.
	Our current method can not handle extreme poses, which we leave as our further work.
	
	\paravspace
	\paragraph{Running time.}
	Tested on an NVIDIA M40 GPU, our face reconstruction, depth estimation and image refinement networks take \textcolor[rgb]{0,0,0}{ $13ms$, $9.5ms$, and $11ms$ respectively to run on one single image. Segmentation takes $15ms$.}
	Interactive portrait manipulation can be easily achieved by our method.

	\section{Conclusion}
	
	We presented a novel approach for single-image 3D portrait modeling, a challenging task that is not properly addressed by existing methods. 
	A new CNN-based 3D head reconstruction pipeline is proposed to learn 3D head geometry effectively without any ground-truth 3D data.
	Extensive experiments and comparisons collectively demonstrated the efficacy of our proposed method for both 3D head reconstruction and single-image pose manipulation. 
	
	\vspace{6pt}
	\noindent\textbf{Acknowledgment.~}{
		This work was partially supported by the National Natural Science Foundation of China under Grants No. 61773062.
	}
	
	{\small
		\bibliographystyle{ieee_fullname}
		\bibliography{deep3dportrait}
	}
 	\clearpage
	

\section*{Supplementary material}

# Deep 3D Portrait from a Single Image
(*Supplementary Material*)

Sicheng Xu[1]*  Jiaolong Yang[2]  Dong Chen[2]  Fang Wen[2]  Yu Deng[3]  Yunde Jia[1]  Xin Tong[2]
[1]Beijing Institute of Technology  [2]Microsoft Research Asia  [3]Tsinghua University

## Abstract

*In this supplementary material, we provide more technical details that are omitted in the main paper due to space constrain. We also present more 3D reconstruction and portrait manipulation results from our method as well as more comparisons with prior art.*

## S1. Network Structure Details

Our depth estimation network uses a ResNet-18 structure [2] as encoder and appends 5 convolution blocks to upsample the feature maps from $8 \times 8$ to the original resolution $256 \times 256$. Each block consists of one transposed conv layer with stride 2 and one conv layer with stride 1. The output channel numbers of the first 4 blocks follow the ResNet-18 encoder at each corresponding resolution. The last transposed conv layer produces 1-channel output (*i.e.*, the depth map). Skip connections are added at $64 \times 64$, $32 \times 32$ and $16 \times 16$ resolutions by concatenating the feature maps from ResNet-18 encoder with the outputs of the corresponding transposed conv layers. In our image refinement module, the structures of $G$ and $D$ networks are identical to [4], except that we change the first conv layer in the residual blocks of $G$ to dilated convolutions with dilation stride 2. Details of their structures can be found in [4].

## S2. More Training Details

**Detailed framework.** In Figure S1, we show the detailed framework of our method as well as the training pipeline for 3D head reconstruction and image refinement.

**Training data for depth estimation network.** In this work, we leverage image pairs to train the hair-ear depth estimation network in a stereo matching setup. Our training image pairs are collected from videos containing human head movements. We select pairs of frames that are temporally close and with small relative rotation (typically within 5 to 15 degrees). Some frame pair examples are presented in Fig. S2.

**Hyperparameter settings.** For the depth estimation network, the weights of $l_{color}$, $l_{grad}$, $l_{face}$, $l_{smooth}$ and $l_{layer}$ are set as 0.4, 0.6, 500 and 100, 100, respectively. Image color is normalized to $[0, 1]^3$. We use batch size of 8, learning rate of $1e-4$ and 63K total iterations. For the image refinement network, we set the weights of $l_{color}(G)$, $l_{prec}(G)$, $l_{adv}(G)$ and $l_{adv}(D)$ as 1, 1, 0.1 and 1, respectively. Image color is normalized to $[-1, 1]^3$ with missing region filled with 0. We use batch size of 32, initial learning rate of $1e-4$ and 200 total epochs. Paired and unpaired images are alternately used in the training iterations. The Adam optimizer [3] is employed to train all these networks.

## S3. More Results and Comparisons

In this section, we present more 3D reconstruction and portrait manipulation results from our method as well as more comparisons with prior art. We also show some pose and expression transfer results from our method.

**3D reconstruction.** Figure S3 presents more 3D head reconstruction results. As can be observed, our method can produce quality 3D geometry for a large variety of face shapes and hair styles.

**Image refinement.** Figure S4 presents more examples of the input and output of the image refinement network. The network can well hallucinate the missing regions (at background, hair, inner mouth and ear) arisen due to head pose and expression change, producing realistic results.

**Pose manipulation.** Figure S5 shows some additional pose manipulation examples from our method. In figure S6, we show more qualitative comparison results with Zhu *et al*. [5] and Chai *et al*. [1].

**Pose and expression transfer.** In this experiment, we estimate the pose and expression of the given source portrait images using our face reconstruction network, and then transfer them to the target subjects using our manipulation pipeline. Figure S7 shows the high-fidelity results from our method for this challenging task.

---

*This work was done when S. Xu was an intern at MSRA.

1

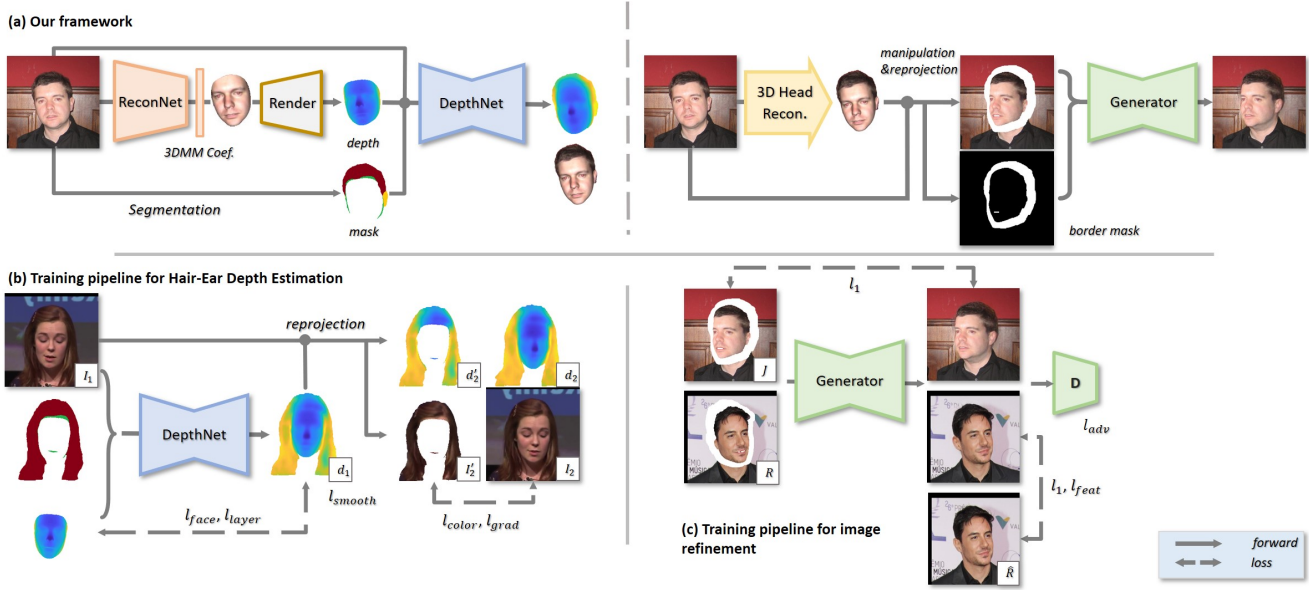

Figure S1: Our framework and training pipelines.

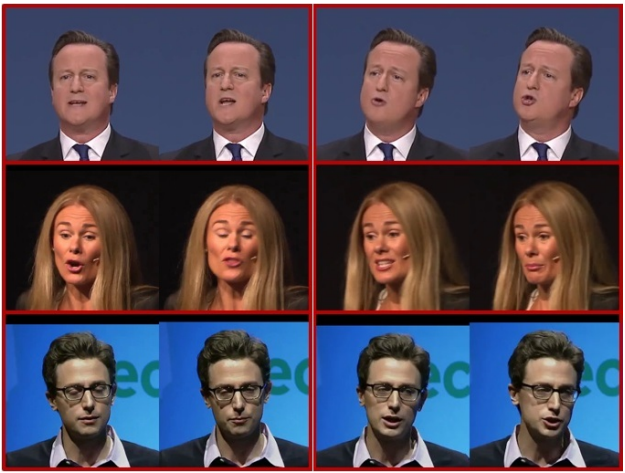

Figure S2: Examples of image pairs used to train the hair-ear depth estimation network. These training image pairs are extracted from videos. The relative rotation angles are within 5 to 15 degrees.

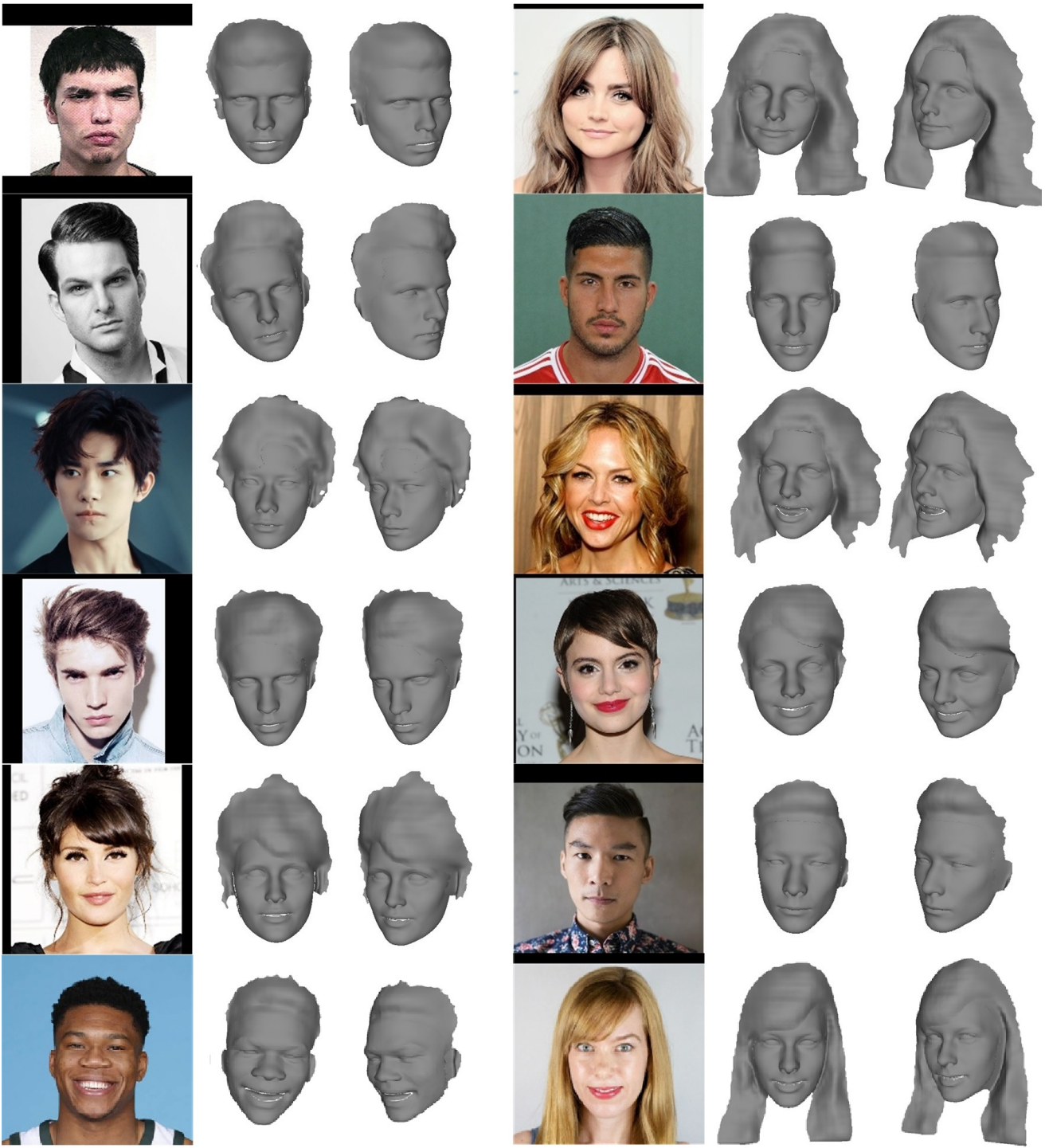

Figure S3: More singe-image 3D head reconstruction results. Our method trained without any ground-truth 3D data can produce quality 3D geometry for a large variety of face shapes and hair styles. Note that all the images (including those in the main paper) are from our test set which are unseen during training.

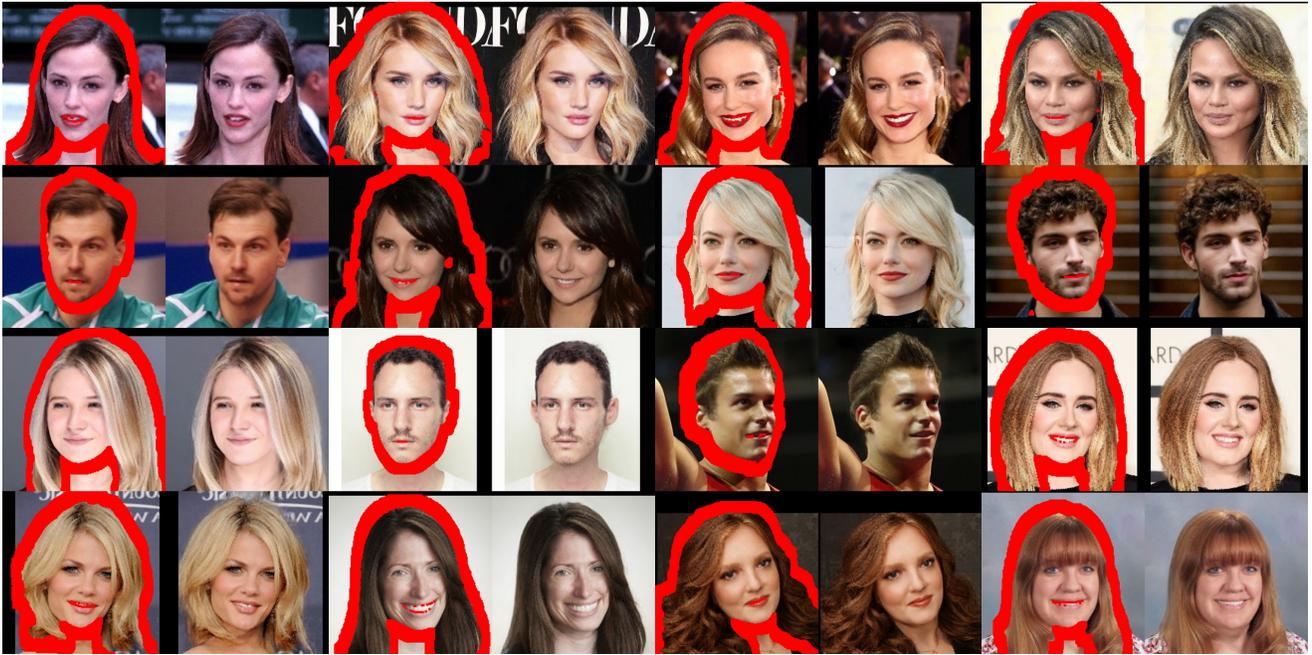

Figure S4: More examples of the input and output of the image refinement network. The missing regions in the inputs images are marked with red. These missing regions, including background, hair, inner mouth and ear, are well hallucinated. Note all the images are from our test sets.

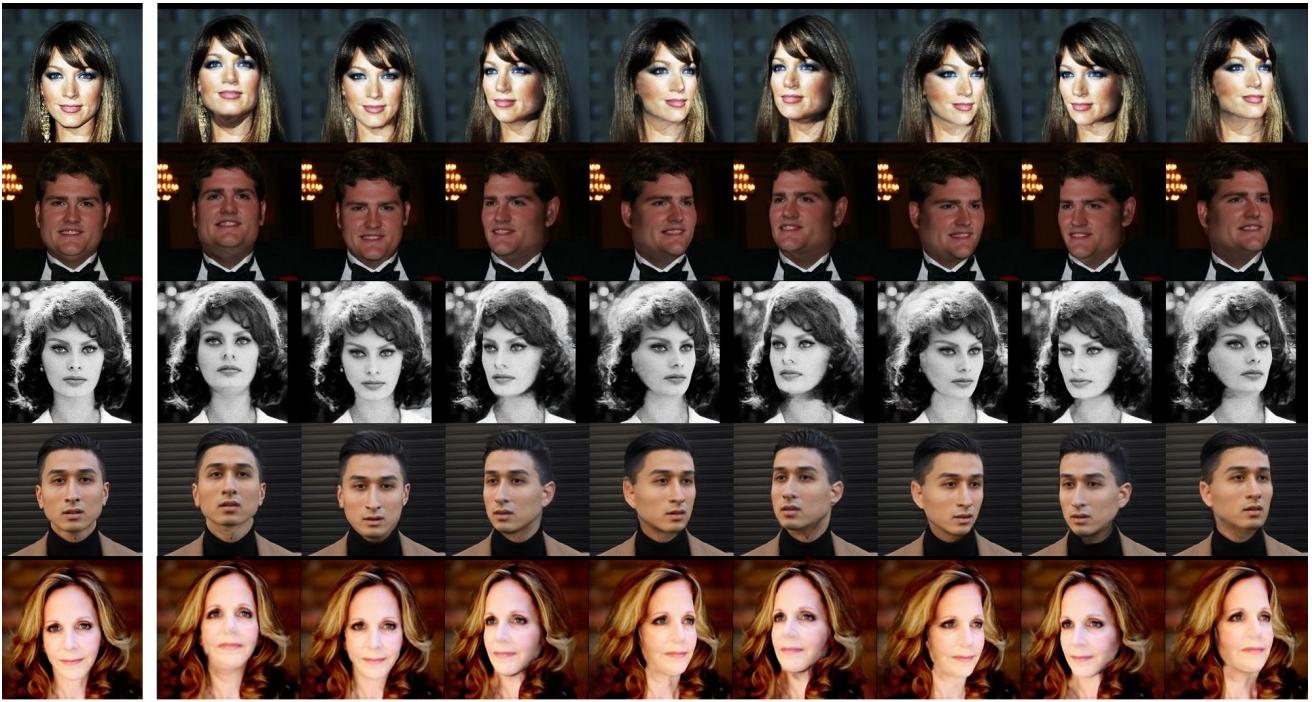

Figure S5: More pose manipulation results. The leftmost column shows the input portrait images, and other columns present the pose manipulation results of our method. Note that all these images are from our test sets.

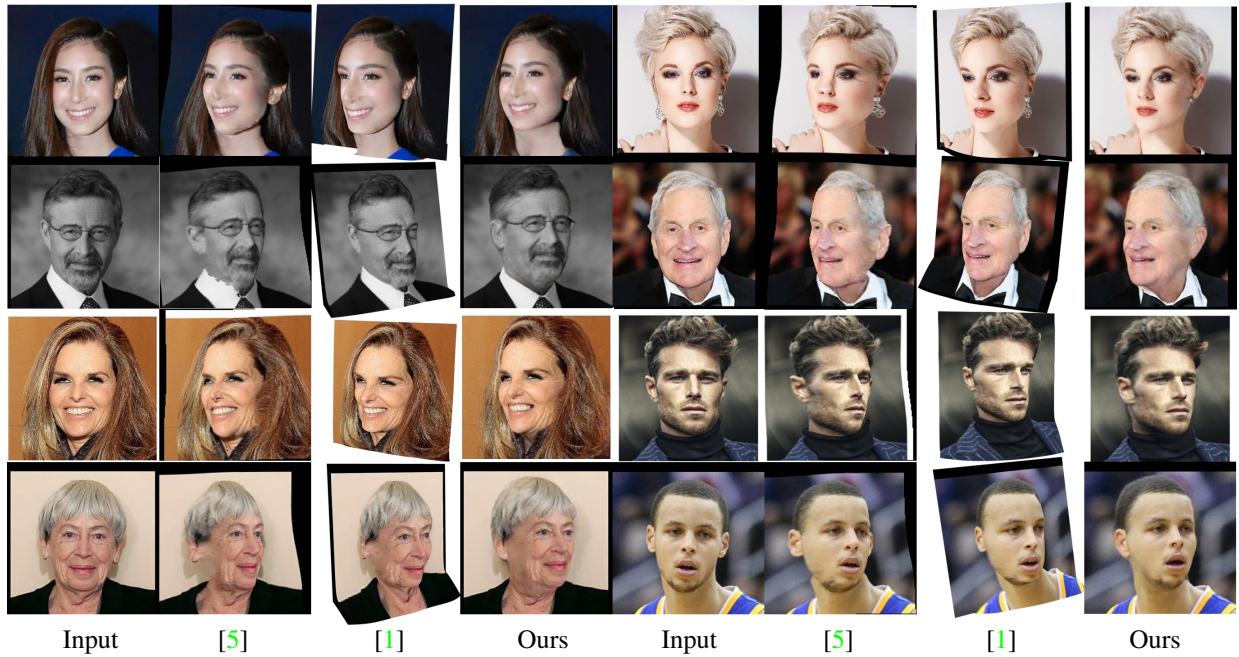

| Input | [5] | [1] | Ours | Input | [5] | [1] | Ours |

Figure S6: More result comparisons with Zhu *et al.* [5] and Chai *et al.* [1].

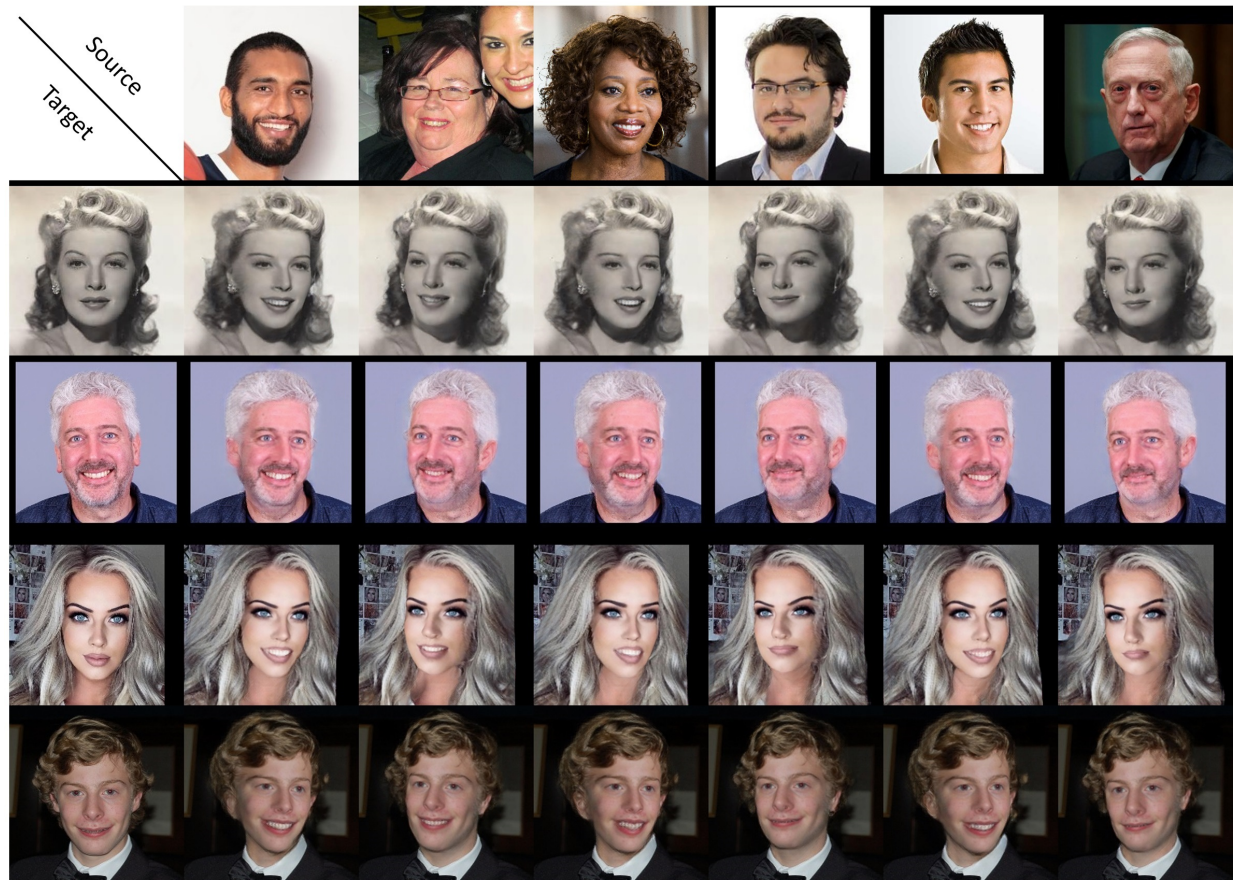

Figure S7: Pose and expression transfer results. The first row and column show the source and target actors, respectively. We transfer the head pose and facial expression from source actors to the target actors in a single-image reenactment setup.



\end{document}